\documentclass[final]{cvpr}

\usepackage{times}
\usepackage{epsfig}
\usepackage{graphicx}
\usepackage{amsmath}
\usepackage{amssymb}

\usepackage[pagebackref=true,breaklinks=true,colorlinks,bookmarks=false]{hyperref}

\def\cvprPaperID{6} 
\def\confYear{CVPR 2021}

\usepackage{silence}
\usepackage[dvipsnames]{xcolor}
\usepackage{multirow}
\usepackage{colortbl}
\usepackage{color}
\usepackage{hyperref}
\usepackage{xspace}
\usepackage{tabularx}

\makeatletter
\@namedef{ver@everyshi.sty}{}
\makeatother
\usepackage{tikz}
\usetikzlibrary{calc,shapes,arrows,backgrounds}
\usepackage{pgfplots}
\pgfplotsset{compat=1.16}
\usepackage{pgfplotstable}
\usepackage{tikzscale}
\usepackage{csvsimple}

\usepackage{siunitx}
\usepackage{booktabs}
\usepackage{makecell}
\usepackage{eqnarray}
\usepackage{bbm}
\usepackage{textcomp}
\usepackage{gensymb}
\usepackage{pifont}
\usepackage{subcaption}
\usepackage{dsfont}
\usepackage{balance}
\usepackage{hyperref}
\usepackage[percent]{overpic}

\usepackage[hyphens]{url}
\usepackage{hyperref}
\usepackage[hyphenbreaks]{breakurl}
\usepackage[pscoord]{eso-pic}

\newcommand{\placetextbox}[3]{
  \setbox0=\hbox{#3}
  \AddToShipoutPictureFG*{
    \put(\LenToUnit{#1\paperwidth},\LenToUnit{#2\paperheight}){\vtop{{\null}\makebox[0pt][c]{#3}}}%
  }%
}%

\newcolumntype{Y}{>{\centering\arraybackslash}X}
\newcolumntype{Z}{>{\raggedleft\arraybackslash}X}

\newcommand*\widebar[1]{%
  \hbox{%
     \vbox{%
      \hrule height 0.5pt 
      \kern0.3ex
      \hbox{%
         \kern-0.1em
         \ensuremath{#1}%
         \kern-0.1em
      }%
     }%
  }%
} 

\begin{document}


\title{Person-MinkUNet: 3D Person Detection with LiDAR Point Cloud}

\author{Dan Jia \hspace{2.5em} Bastian Leibe\\

Visual Computing Institute, RWTH Aachen\\
{\tt\small \{jia, leibe\}@vision.rwth-aachen.de}
}

\placetextbox{0.27}{0.98}{accepted as an extended abstract in JRDB-ACT Workshop at CVPR21}

\maketitle
\section{Introduction}
Existing methods for processing point clouds can largely be grouped into two categories: those that directly operates on points~\cite{Qi17CVPR,Qi17NIPS,Thomas19ICCV}, and those that relies on voxelization~\cite{Zhou17CVPR,Graham18CVPR,Yan18Sensors,Choy19CVPR,Lang19CVPR}.
Of the latter category, submanifold sparse convolution~\cite{Graham18CVPR,Choy19CVPR} has gained success on various benchmarks, including ScanNet~\cite{Dai17CVPR} and SemanticKITTI~\cite{Behley19ICCV}.
For the task of 3D object detection, however, the current state-of-the-art is established by CenterPoint~\cite{Yin21CVPR} with VoxelNet~\cite{Zhou17CVPR} backbone. 

In this preliminary work we attempt to apply submanifold sparse convolution to the task of 3D person detection.
In particular, we present Person-MinkUNet, a single-stage 3D person detection network based on Minkowski Engine~\cite{Choy19CVPR} with U-Net~\cite{Ronneberger15MICCAI} architecture.
The network achieves a 76.4\% average precision (AP) on the JRDB 3D detection benchmark~\cite{Martin21PAMI}.\footnote{
Benchmark entry: \url{https://jrdb.stanford.edu/leaderboards/results/553}}

\section{Person-MinkUNet}

The input to Person-MinkUNet is voxelized point cloud.
In this work, we used voxel size (0.05m, 0.05m, 0.1m).
A backbone network, implementation taken from~\cite{Tang20ECCV}, is used to extract features for each non-empty voxels.
It is a submanifold sparse convolution network with ResNet20~\cite{He16CVPR} architecture and U-Net~\cite{Ronneberger15MICCAI} connections.
A fully connected layer is then used to regress 3D bounding boxes from the extracted features.
These box proposals, after non-maximum suppression, are directly used as detections, with no refinement stage.

Each bounding box is described with 7 box parameters (location, dimension, z-rotation) and 1 classification score.
To predict these values, we use the following parametrization.
For assigning the classification target, we use
\begin{equation}
    t_c =
    \begin{cases}
      1.0 - \frac{d}{0.5 \times (D + \epsilon)} & \text{if voxel in box}\\
      0 & \text{otherwise}
    \end{cases} 
\end{equation}
where $d$ is the distance from the voxel center to the box center, and $D=\sqrt{l^{2}+w^{2}+h^{2}}$ is the box diagonal length, and $\epsilon=0.2$ is a hyper-parameter.
For the dimensions, we use
\begin{equation}
    t_l = \log{(l / \bar{l})}, t_w = \log{(w / \bar{w})}, t_h = \log{(h / \bar{h})}
\end{equation}
where $\bar{l}=0.9$, $\bar{w}=0.5$, $\bar{h}=1.7$ are the average box length, width, and height in the training set.
For the z-rotation $\phi$, we use the bin-based regression from~\cite{Shi19CVPR} with twelve non-overlapping bins.

We use cross-entropy loss for classification and $L$2 loss for regression.
The total loss is given as
\begin{align}
    L_{total} &= L_{cls} + L_{box} + 0.1 \times L_{\phi} \\
    L_{box} &= L_{xyz, reg} + L_{lwh, reg} \\
    L_{\phi} &= L_{\phi, cls} + 0.1 \times L_{\phi, reg}
\end{align}
Box and z-rotation is only supervised when the voxel falls within a bounding box.

\section{Evaluation}

We use the joint point cloud from the upper and lower velodyne LiDARs.
We train the network for 40 epochs with Adam optimizer~\cite{Kingma15ICLR}, using an initial learning rate 10$^{-3}$ for the first 15 epochs, and exponentially decay to 10$^{-6}$.
Random scaling between (0.95, 1.05) and random rotation along the vertical axis is used for data augmentation.

For baseline, we train a CenterPoint using the original training setup from~\cite{Yin21CVPR}, except we reduce the voxel size to (0.05m, 0.05m, 0.2m) and disable the cut-paste data augmentation.

On the JRDB validation set, Person-MinkUNet scores 71.5\%~AP, and the baseline CenterPoint scores 66.0\%~AP.\footnote{
Qualitative results can be found at \url{https://youtu.be/RnGnONoX9cU}}

\section{Benchmark Submission}
The benchmark submission is trained using both training and validation set.
The result was obtained with a single model with no ensembles.

{\small
\bibliographystyle{ieee_fullname}
\bibliography{abbrev_short,mybib}

\begin{thebibliography}{10}\itemsep=-1pt

\bibitem{Behley19ICCV}
Jens Behley, Martin Garbade, Andres Milioto, Jan Quenzel, Sven Behnke, Cyrill
  Stachniss, and Juergen Gall.
\newblock {SemanticKITTI: A Dataset for Semantic Scene Understanding of LiDAR
  Sequences}.
\newblock In {\em ICCV}, 2019.

\bibitem{Choy19CVPR}
Christopher Choy, JunYoung Gwak, and Silvio Savarese.
\newblock {4D Spatio-Temporal ConvNets: Minkowski Convolutional Neural
  Networks}.
\newblock In {\em CVPR}, 2019.

\bibitem{Dai17CVPR}
Angela Dai, Angel~X. Chang, Manolis Savva, Maciej Halber, Thomas Funkhouser,
  and Matthias Nie{\ss}ner.
\newblock {ScanNet: Richly-annotated 3D Reconstructions of Indoor Scenes}.
\newblock In {\em CVPR}, 2017.

\bibitem{Graham18CVPR}
Benjamin Graham, Martin Engelcke, and Laurens van~der Maaten.
\newblock {3D Semantic Segmentation with Submanifold Sparse Convolutional
  Networks}.
\newblock In {\em CVPR}, 2018.

\bibitem{He16CVPR}
Kaiming He, Xiangyu Zhang, Shaoqing Ren, and Jian Sun.
\newblock {Deep Residual Learning for Image Recognition}.
\newblock In {\em CVPR}, 2016.

\bibitem{Kingma15ICLR}
Diederik~P. Kingma and Jimmy Ba.
\newblock {Adam: A Method for Stochastic Optimization}.
\newblock In {\em ICLR}, 2015.

\bibitem{Lang19CVPR}
Alex~H. Lang, Sourabh Vora, Holger Caesar, Lubing Zhou, Jiong Yang, and Oscar
  Beijbom.
\newblock {PointPillars: Fast Encoders for Object Detection From Point Clouds}.
\newblock In {\em CVPR}, 2019.

\bibitem{Martin21PAMI}
Roberto Martin-Martin*, Mihir Patel*, Hamid Rezatofighi*, Abhijeet Shenoi,
  JunYoung Gwak, Eric Frankel, Amir Sadeghian, and Silvio Savarese.
\newblock {JRDB: A Dataset and Benchmark for Visual Perception for Navigation
  in Human Environments}.
\newblock {\em PAMI}, 2021.

\bibitem{Qi17CVPR}
Charles~R Qi, Hao Su, Kaichun Mo, and Leonidas~J Guibas.
\newblock {PointNet: Deep Learning on Point Sets for 3D Classification and
  Segmentation}.
\newblock In {\em CVPR}, 2017.

\bibitem{Qi17NIPS}
Charles~Ruizhongtai Qi, Li Yi, Hao Su, and Leonidas~J. Guibas.
\newblock {PointNet++: Deep Hierarchical Feature Learning on Point Sets in a
  Metric Space}.
\newblock In {\em NIPS}, 2017.

\bibitem{Ronneberger15MICCAI}
Olaf Ronneberger, Philipp Fischer, and Thomas Brox.
\newblock {U-Net: Convolutional Networks for Biomedical Image Segmentation}.
\newblock {\em Medical Image Computing and Computer-Assisted Intervention
  (MICCAI)}, 9351:234--241, 2015.

\bibitem{Shi19CVPR}
Shaoshuai Shi, Xiaogang Wang, and Hongsheng Li.
\newblock {PointRCNN: 3D Object Proposal Generation and Detection From Point
  Cloud}.
\newblock In {\em CVPR}, 2019.

\bibitem{Tang20ECCV}
Haotian* Tang, Zhijian* Liu, Shengyu Zhao, Yujun Lin, Ji Lin, Hanrui Wang, and
  Song Han.
\newblock {Searching Efficient 3D Architectures with Sparse Point-Voxel
  Convolution}.
\newblock In {\em ECCV}, 2020.

\bibitem{Thomas19ICCV}
H. Thomas, C. Qi, Jean-Emmanuel Deschaud, B. Marcotegui, F. Goulette, and L.
  Guibas.
\newblock {KPConv: Flexible and Deformable Convolution for Point Clouds}.
\newblock In {\em ICCV}, 2019.

\bibitem{Yan18Sensors}
Bo~Li Yan~Yan, Yuxing~Mao.
\newblock {SECOND: Sparsely Embedded Convolutional Detection}.
\newblock {\em Sensors}, 2018.

\bibitem{Yin21CVPR}
Tianwei Yin, Xingyi Zhou, and Philipp Kr{\"a}henb{\"u}hl.
\newblock {Center-based 3D Object Detection and Tracking}.
\newblock In {\em CVPR}, 2021.

\bibitem{Zhou17CVPR}
Yin Zhou and Oncel Tuzel.
\newblock {VoxelNet: End-to-End Learning for Point Cloud Based 3D Object
  Detection}.
\newblock In {\em CVPR}, 2017.

\end{thebibliography}
}

\end{document}